\documentclass[sigconf]{acmart}

\usepackage{graphicx}
\usepackage{caption}

\AtBeginDocument{%
  \providecommand\BibTeX{{%
    \normalfont B\kern-0.5em{\scshape i\kern-0.25em b}\kern-0.8em\TeX}}}

\copyrightyear{2024}
\acmYear{2024}
\setcopyright{acmlicensed}\acmConference[GoodIT '24]{International Conference on Information Technology for Social Good}{September 4--6, 2024}{Bremen, Germany}
\acmBooktitle{International Conference on Information Technology for Social Good (GoodIT '24), September 4--6, 2024, Bremen, Germany}
\acmDOI{10.1145/3677525.3678667}
\acmISBN{979-8-4007-1094-0/24/09}

\acmConference[GoodIT'24]{International Conference on Information Technology for Social Good}{September 4--6,
  2024}{Bremen, Germany}
\acmBooktitle{GoodIT'24: International Conference on Information Technology for Social Good, 4--6 September 2024, Bremen, Germany} 
\acmISBN{978-1-4503-XXXX-X/18/06}

\begin{document}

\title{Quantitative Information Extraction from Humanitarian Documents}


\author{Daniele Liberatore}
\affiliation{%
  \institution{ISI Foundation}
  \streetaddress{Via Chisola 5}
  \city{Turin}
  \country{Italy}
  \postcode{10126}
}

\author{Kyriaki Kalimeri}
\orcid{0000-0001-8068-5916}
\affiliation{%
  \institution{ISI Foundation}
  \streetaddress{Via Chisola 5}
  \city{Turin}
  \country{Italy}
  \postcode{10126}}
\email{kyriaki.kalimeri@isi.it}

\author{Derya Sever}
\affiliation{%
  \institution{Data Friendly Space}
  \streetaddress{205 John Wythe Place}
  \city{Williamsburg}
  \state{VA}
  \country{USA}
  \postcode{23185}}
\email{derya@datafriendlyspace.org}

\author{Yelena Mejova}
\orcid{0000-0001-5560-4109}
\affiliation{%
  \institution{ISI Foundation}
  \streetaddress{Via Chisola 5}
  \city{Turin}
  \country{Italy}
  \postcode{10126}}
\email{yelenamejova@acm.org}

\renewcommand{\shortauthors}{Liberatore, et al.}

\begin{abstract}

Humanitarian action is accompanied by a mass of reports, summaries, news, and other documents. 
To guide its activities, important information must be quickly extracted from such free-text resources.
Quantities, such as the number of people affected, amount of aid distributed, or the extent of infrastructure damage, are central to emergency response and anticipatory action.
In this work, we contribute an annotated dataset for the humanitarian domain for the extraction of such quantitative information, along side its important context, including units it refers to, any modifiers, and the relevant event.
Further, we develop a custom Natural Language Processing pipeline to extract the quantities alongside their units, and evaluate it in comparison to baseline and recent literature.
The proposed model achieves a consistent improvement in the performance, especially in the documents pertaining to the Dominican Republic and select African countries.
We make the dataset and code available to the research community to continue the improvement of NLP tools for the humanitarian domain.

\end{abstract}




\maketitle

\section{Introduction}

Disasters induced by human activities, such as conflicts and climate change, as well as natural disasters, are causing displacement and urgent needs for healthcare and supplies for millions. 
In emergencies, the initial 72 hours are crucial for saving lives.
Much of the humanitarian information for emergency response and anticipatory action is presented in ``secondary'' data sources, including reports, assessments, news, and other textual forms. These documents are fundamental for identifying vulnerable groups, assessing needs and response gaps, and deciding on the type of relief activities to be undertaken during and before a crisis.
The analysis of these data rely heavily on extracting key information and organizing it according to pre-defined, domain-specific structures and guidelines known as Humanitarian Analysis Frameworks.\footnote{\url{https://2021.gho.unocha.org/delivering-better/joint-intersectoral-analysis-framework/}} This organization helps guide decisions on the impact, needs, and resource allocation to assist vulnerable communities effectively and timely.
However, analyzing secondary data within the critical 24-72 hour window post-crisis is challenging due to time and resource constraints. 

Natural Language Processing (NLP) is advancing traditional practices in the humanitarian field, supporting, among other tasks, the extraction of information from unstructured data. 
Platforms like the Data Entry and Exploration Platform (DEEP)\footnote{\url{https://thedeep.io/}} offer tools for compiling, storing, and structuring data and information.
However, the capabilities of current platforms are limited to classifying qualitative data based on analytical frameworks. There is a lack of open models for extracting quantitative information from text for deeper and faster analysis. While some explorations have been done using closed-source large language models (LLMs) during emergencies, their cost and the opaqueness of their training datasets limit their utility for evidence-based humanitarian action~\cite{peng2024}.
New initiatives like the Joint Analysis Workspace (JAWS)\footnote{\url{https://www.datafriendlyspace.org/our-work/deep}} aim to integrate qualitative and quantitative data using advanced technologies to elicit expert judgment, highlighting the need for developing open-source models for quantitative information extraction from unstructured text with traceability.


In this work, we propose a methodology for quantity extraction and contribute a richly annotated dataset of humanitarian documents to evaluate this method. 
In collaboration with humanitarian partners, we develop an annotation schema that captures the number and its modifiers, units, and the broader event context. 
This schema was applied by expert coders to a large collection of humanitarian crisis reports~\cite{fekih-etal-2022-humset}, resulting in a dataset of 755 documents containing 4,352 detailed quantity annotations. 

We evaluate our model against established NLP tools (Spacy)\footnote{\url{https://spacy.io/}} and a recent quantity extractor from the literature~\cite{almasian-etal-2023-cqe}, demonstrating improvements in precision and recall for extracting numbers and their associated units. 
The performance was notably high in documents related to the humanitarian sectors of \emph{shelter} and \emph{WASH} (\emph{water, sanitation and hygiene}), though further enhancements are needed for the \emph{nutrition}, \emph{food security}, and \emph{livelihoods} categories. 
We have made the annotated dataset and system code available\footnote{\url{https://github.com/dani-libe/HumQuant}} to ensure reproducibility and to encourage further development of domain-specific NLP tools for the humanitarian sector.





\section{Related Works}

As frontier data are increasingly more employed in the humanitarian response, data and method biases assessment becomes crucial ~\cite{sekara2024opportunities,sartirano2023strengths,beiro2022fairness}. 

\subsection{Quantity Extraction}

The ambition of this work is to contribute to the development of a system similar to the ``Discrete Reasoning Over the text in the Paragraph'' proposed by Dua et al.~\cite{duaDROPReadingComprehension2019}, which may involve quantitative reasoning tasks such as sorting, counting, and basic arithmetic on an unstructured text input. The authors created a benchmark dataset for the task using passages from Wikipedia and had crowdworkers craft ``challenging'' questions. 
Although such general-purpose applications are thus far intractable (at the time, the best system achieved F1 0.32 on the generalized accuracy metric), an intermediate step is the extraction and summarization of quantitative data. 
The process begins by the identifying the span of text relevant to a numeral (a ``span identification task'') \cite{gopfertMeasurementExtractionNatural2022}, followed by a unit or modifier identification. 
Number extraction is relevant to a wide variety of tasks, including mathematical problem solving \cite{zhang2020gap} and equation parsing \cite{roy-etal-2016-equation}.
These can be then inputs to steps that provide more semantically rich relations, such as those achieved by the numerical relation extraction operation, which may provide associations such as ``inflation\_rate(India, 10.9\%)'' \cite{sarawagiNumericalRelationExtraction}.
To bolster cooperation and standardize the benchmarks, competitions such as MeasEval \cite{harper-etal-2021-semeval} have been organized by the NLP community.
Specifically, the task concerns extracting counts, measurements, and related context from scientific documents in the aim of supporting the automatic creation of Knowledge Graphs from unstructured text.
Most approaches use a ``cascaded approach'', in which the output of earlier stage (identifying the numerals in text) is used in the latter stage (identifying units or other contextual features) \cite{davletov-etal-2021-liori-semeval}.
The system proposed in this work uses a similarly structured pipeline, which can be extended in the future work.

Several domain-specific quantitative information retrieval systems have been proposed.
For instance, Yang et al.~\cite{yangNaturalLanguageProcessing2020} built a rule-based system to extract information from clinical notes about patients who have undergone CT scanning for lung cancer. 
Trained on 200 documents, their system achieves F1 score of 0.95 in the strict setting and 0.96 in the lenient one.
Chen et al.~\cite{chenNumeralAttachmentAuxiliary2019} developed a dataset for ``numeral attachment'' task called NumAttach, which consists of 7984 tweets relevant to finance. An expert from the trading desk of a commercial bank annotated them for numerals and potential reasons why it was mentioned, such as ``asset'' or ``liability''. The authors use character embeddings to capture out-of-vocabulary (OOV) information, and to capture separately the context preceding and following context.
More recently, Almasian et al.~\cite{almasianCQEComprehensiveQuantity2023} propose a rule-based quantity extraction framework Comprehensive Quantity Extractor (CQE) that extracts values and units from text, and performs normalization and standardization of units. They test the system on a collection of news articles spanning topics such as economics, sports, technology, and cars. 
The performance of their model achieves F1 score in the range of 0.79 - 0.93 on number and unit extraction tasks in the various subtasks.
Being some of the latest and generally-applicable proposed models, we compare our proposed model to CQE\footnote{\url{https://github.com/vivkaz/CQE/tree/main}}. 
Finally, number detection has been a feature of off-the-shelf state-of-the-art tools such as spaCy\footnote{\url{https://spacy.io/usage/linguistic-features}}, that detect numbers as a part of the part-of-speech (PoS) tagging. 
Trained on large datasets of general language use, including the large lexical database WordNet and corpora such as OntoNotes spanning various genres of text\footnote{\url{https://spacy.io/models/en}}, the model achieves training performance of F1 0.73 for the Quantity label and F1 0.81 for Ordinal.
We also use spaCy as another baseline, as well as a component of our model's pipeline.

Although MeasEval supports the development of many systems (75 submissions from 25 participants were submitted in 2021), the selection of the dataset constrains their application to the scientific writing.
Generalization challenges remain, especially in the light of domain-specific peculiarities of different settings.
For instance, clinical reports may contain a variety of numerical information regarding the attributes of the patient, test results, outcome estimates, etc.~\cite{hanauer2019complexities}, whereas some technical documents may contain tables and schematics containing quantitative information \cite{opasjumruskit2021automatic}. 
Also, some documents are only available as image or PDF files, necessitating an Optical Character Recognition (OCR) pre-processing. 
Further, domain knowledge is necessary to interpret and select quantities relevant to the tasks at hand.


\subsection{Data Extraction for Humanitarian Domain}

In the humanitarian domain, extracting quantitative information from text is crucial for crisis response and decision-making. 
However, traditional methods struggle to rapidly process open-ended responses; \cite{kreutzer2020} proposed NLP as a solution to analyze large sets of qualitative responses effectively, discussing the potential of NLP to transform humanitarian needs assessments by providing more nuanced insights from qualitative data collected during crises.  
This approach focused on potentially improving the speed and accuracy of operational decisions in humanitarian responses by enabling the extraction of detailed information from the affected population’s voice responses.

The sensitive nature of the humanitarian domain calls for approaches that minimise the classification errors and the uncertainty of the decision making process. Hence, early NLP tools developed for humanitarian support include rule based systems that ensure explainability of the results, such as Marve, a system for extracting measurement values from plain text, which was motivated by the advances by NASA’s Hyperspectral Infrared Imager (HyspIRI) mission and its scientific reports \cite{hundmanMeasurementContextExtraction2017}.

Focusing on news articles, \cite{ning2022}, proposed a framework that identifies quantities along with their type, time, and location from unstructured text. The ability to effectively parse and understand these spatiotemporal aspects from textual data can significantly enhance quantitative reporting and situational analysis in humanitarian contexts. 
News articles, though, are carefully edited documents;  humanitarian reports on the other hand, which are the focus of this study, are often written during emergencies, and hence are more dense and complex documents.

In a recent study, \cite{peng2024} explored the effectiveness of distilling a small language model (LM) from large language models (LLMs) for various information extraction tasks. Their study primarily addresses general information extraction tasks beyond the humanitarian context, however, the methodology offers valuable insights into adapting large-scale NLP models to more focused and resource-efficient applications.
As pointed out by the authors, the model suffers from inherit biases, hence, its direct adaptation to high-risk settings should be avoided. 



\section{Data Collection}

The present work builds on one of the largest humanitarian datasets: the HumSet \cite{fekih-etal-2022-humset}; a collection of excerpts of humanitarian crisis reports from various sources (including news articles and on-ground assessments) from across the globe published in the time span of 2018 to 2021. 
HumSet synthesizes 11 different frameworks into a unified Humanitarian Analysis Framework,  including annotated excerpts with coarse and fine-grained categories that represent its main subjects.
The metadata also include the document's language and the country code of the location report concerns. 

In this study, we select entries in the English language, comprising of 90534 unique entries. Out of those, we select those with numeric information, and more specifically, we select those that contain at least 3 numbers detected by the baseline system (see Section 5.1).
In brief, the baseline system is based on the Spacy \textit{en\_core\_web\_sm} model and  a set of named entity recognizer rules both to avoid the identification of numbers in which we are not interested in (e.g.~dates) and to consider, instead, all those written in standardized ways (e.g.~different styles of writing percentages).
After identifying excerpts having at least 3 numbers, we apply additional filtering conditions: the entry must have the country code and sector information and should be at least 100 tokens long. 
The final dataset employed in this study contains 2872 unique texts out of the initial 90534.

\begin{figure}
    \begin{minipage}[t]{0.75\linewidth}
    \centering
    \includegraphics[width=0.75\textwidth]{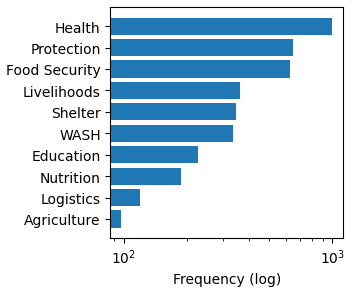}
    \end{minipage}
    \begin{minipage}[t]{0.95\linewidth}
    \centering
    \includegraphics[width=\textwidth]{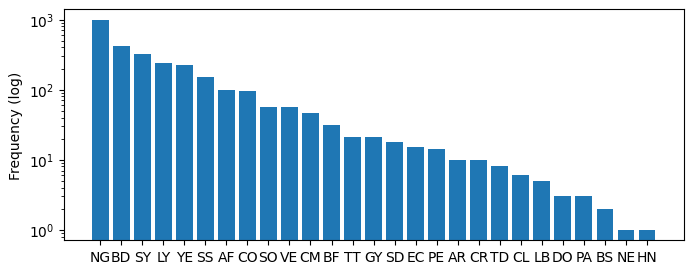}
    
    \end{minipage}
    \caption{Distribution over sectors and countries of the selected excerpts from HumSet.}
    \Description{Distribution over sectors and countries of the selected excerpts from HumSet.}
    \label{fig:some_dists}
\end{figure}

Finally, to ensure a representative sample of excerpts for the annotation we stratify over the sector and country for each entry with the constraint of having three samples for each sector-country pair (see Fig.~\ref{fig:some_dists}).
Since some pairs are present less often than three times, in such cases we take the total number of entries. When the amount of excerpts extracted in this way does not reach the desired number of entries, we sample the remaining ones in the randomly, without any stratification. 
In Figure \ref{fig:some_dists} we show the distribution of two of the main features, \textit{sector} and \textit{country code}, in this final set of 780 unique excerpts. 

\section{Humanitarian-focused Information Annotation}

The selected excerpts are then annotated by three domain experts on the semantic annotation platform INCEpTION\footnote{\href{https://inception-project.github.io/}{https://inception-project.github.io/}} which allows the annotators to tag not only tokens, but also relations between them.
The annotation consists of two rounds: the first of 90 excerpts (20 for each person plus 10 for each annotator pair) and the second one of 690 entries (200 for each person plus 30 for each annotator pair). We compute the annotator agreement on the common excerpts each time.
Between the two rounds a Q\&A session took place in order to answer the annotator doubts and to improve the quality of the second, and bigger, annotation round.

\begin{figure*}
    \centering
    \includegraphics[width=0.9\textwidth]{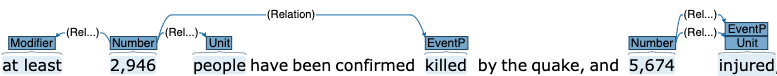}
    \caption{Annotation example. The annotation centers around the Number label, while the Unit, and Events are identified in the relation to the Number.}
    \Description{Annotation example. The annotation centers around the Number label, while the Unit, and Events are identified in the relation to the Number.}
    \label{fig:annotation-example}
\end{figure*}

After several refinements in consultation with the domain experts, the final version of the annotation schema contains six labels that represent all of the information that capture the definition of quantity in humanitarian context:
\textit{Number} -- a numeral written in digits or letters, \textit{Unit} --  entities (people or objects) related to a specific Number, \textit{Modifier} -- any word that changes the degree of uncertainty of a certain Number, and three event tags including \textit{EventP} -- that directly impact people, \textit{EventA} -- related to the assistance given to people in need, and \textit{EventO} -- any other kind of event not described by the last two categories. 
The separation of event tags into three has been suggested by the experts from DEEP, such that the events annotated in the data could be more easily related to the pillar/sub-pillar Humanitarian Analysis Framework used by the system \cite{fekih-etal-2022-humset}.
The schema,\footnote{Available at \url{https://github.com/dani-libe/HumQuant/blob/main/annotation_schema.pdf}} describes in more detail the definitions and special cases of each of these labels.
Crucially, the annotation centers around the number label -- it is annotated first, and others are identified in the relation to the number (that is, if no number has been annotated, the others are also not annotated).
For each label, we provided the annotators with examples of texts with correct annotations.
Also, the experts suggested that we constrain the kinds of numbers which are of interest in this task by excluding dates, sections and titles, and page numbers. 
An example labeled sentence is shown in Figure \ref{fig:annotation-example}.
The annotators reported to have fully annotated an average of 8 excerpts per hour.

To measure the extent to which the annotators understood and complied with the guidelines, we compute an inter-annotator agreement score over both rounds to assess the disagreement between the three annotators.
Since the task deals with the character-level selection of text we use two metrics.
First, we use a standard metric for annotator agreement for tasks such as Named Entity Recognition, the $\gamma$ coefficient, as it captures both the token locating and labeling, and handles partially overlapping selections \cite{mathet-etal-2015-unified}.
Secondly, we define a metrics that mimics a classification task. 
Given an annotator pair, we consider one annotator as ground truth and compute precision, recall, and $F_1$ measures for the second.
Two spans of text can be said to match in two ways: (1) if they overlap exactly, or (2) if they overlap with at least one character.
Both metrics are computed on each token label (for the events we consider EventP, EventA and EventO as a generic Event) and at the aggregate level, by considering all the labels together.

\begin{table}
\parbox{\linewidth}{
\centering
    \begin{tabular}{l|l|l|l|l}\toprule
& \multicolumn{2}{c|}{First round} & \multicolumn{2}{c}{Second round} \\
& $F_1$ & $\gamma$ & $F_1$  & $\gamma$ \\ \midrule
Number      & 0.89  & 0.95  & 0.94  & 0.93 \\
Unit        & 0.81  & 0.81  & 0.86  & 0.85 \\
Modifier    & 0.80  & 0.81  & 0.68  & 0.80 \\
Event (all) & 0.58  & 0.62  & 0.75  & 0.65 \\ \midrule
Aggregated  & 0.80  & 0.82  & 0.86  & 0.80 \\ \bottomrule
    \end{tabular}
    \caption{Inter-annotator agreement scores, $F_1$ and $\gamma$ coefficient, for first (development) and second (final) rounds of annotation.}
    \label{tab:fr-agreement}
}
\vfill
\parbox{\linewidth}{
\centering
    \begin{tabular}{l|r}\toprule
             & Frequency      \\ \midrule
    Number   & 4352           \\
    Unit     & 3011           \\
    Modifier & 461            \\
    EventP   & 812            \\
    EventA   & 437            \\
    EventO   & 1244           \\ \bottomrule
    \end{tabular}
    \caption{Distribution of labels in the annotated dataset (in both rounds).}
    \label{tab:labels-dist}
    \vspace{-0.5cm}
}
\end{table}

Table \ref{tab:fr-agreement} shows the agreement scores computed on the first and second round. 
We find that the scores for the Number and Unit labels in both rounds are quite high, probably due to their straightforward definition. 
The same applies for the Modifier label, even if, in the second round our custom score is lower than the $\gamma$ one. This is due to the fact that when computing the $\gamma$ score, we discard all the excerpts which do not contain at least one annotation for the label of interest by each annotator, while, in our custom metric, we heavily penalize these documents by setting precision and recall to 0.
The different scores for the general Event label, instead, could be related to the way we consider overlapping spans. In our metric we do not take in account the size of the shared span of text between two annotations, while this is relevant for the computation of the $\gamma$ score.
Moreover, we find that the Q\&A session that occurred between the two rounds has been useful especially for a better comprehension of the Event labels.

\begin{figure}
\begin{minipage}[b]{0.8\linewidth}
    \centering
    \includegraphics[width=\textwidth]{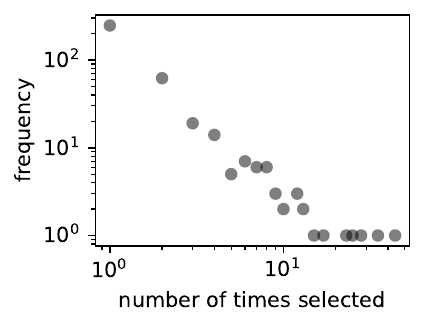}
    \captionof{figure}{Frequencies of units.}
    \Description{Frequencies of units.}
    \label{fig:unit_freq}
\end{minipage}
\vspace{0.5cm}

\begin{minipage}[b]{\linewidth}
\footnotesize
\centering
    \begin{tabular}{llll}
households (44) & individuals (17) & ngos (12) & girls (9) \\
children (35) & years (15) & suspected measles  & proportion of \\
people (28) & m-o-m (13) & \hspace{0.3cm}cases (12) &  \hspace{0.2cm}households (9)\\
cases (25) & beds (13) & deaths (10) & death (8) \\
syp (23) & hand pumps (12) & students (10) & schools (8) \\
 &  & women (9) & systems (8) \\
    \end{tabular}
    \captionof{table}{Top 20 units in annotated data (frequency).}
    \label{tab:units_top}
\end{minipage}
\end{figure}

In the final dataset, 755 excerpts (out of the total of 780) have at least one token annotated.
In the case of the excerpts that were used for inter-annotator agreement, the final annotations were selected randomly (from one or another labeler).
Table \ref{tab:labels-dist} shows the number of tokens (note they can span several words) labeled with each category of label.
As expected, the Number label is the most represented because during the data sampling for the annotation phase we favored excerpts that are likely to contain one.
Despite different frequencies of number and unit labels, 11\% of the numbers are not accompanied by a unit, because often several numbers are referring to the same unit token. Also, 11\% of the extracted numbers are accompanied by a modifier.
Among the events, the most frequently selected was EventO (``other'' event), followed by EventP (that affecting people) and EventA (concerning some assistance); 15\% of numbers was not associated with an accompanying event label.
Finally, Figure \ref{fig:unit_freq} shows the frequency distribution of units, and Table \ref{tab:units_top} shows the top 20 units.
Most units (64.5\%) are selected only once in our dataset, pointing to a great variety of subjects present. 
Out of the most standard ones, we find ``households'', ``children'', ``people'', and ``cases''. 
We also find currencies (``syp'' stands for the Syrian Pound), objects (``beds''), and even words indicating events (``deaths'') and time (``m-o-m'' means ``month-on-month'').
We make available this dataset as a resource to the community.\footnote{\url{https://github.com/dani-libe/HumQuant}}

\section{Quantity Extraction System Design}

\subsection{Number extraction}

Our approach for the extraction of quantitative data from text is based on a multistage pipeline. At first we identify only the raw numerals and then we feed them to the second module whose goal is the extraction of the related unit (if any).
The number extractor is based on the named entity recognition and part of speech tagging modules from the "en\_core\_web\_sm" model provided by the Spacy library.
This approach fits our needs well, since our aim is to maximize the chance of identifying any kind of relevant number (that could be written in digits or in letters). 
We use the Spacy library, due to its good speed/performance trade-off and also because it is the only relevant NLP library that offers a wide range of labels for the NER task.
For our task, we select NER labels related to the numbers of our interest, specifically \emph{cardinal}, \emph{quantity}, \emph{money}, and \emph{percent}.
Moreover, we are able to integrate matching rules, using the EntityRuler component offered by Spacy that allows us to ensure the detection of both relevant and non-relevant numbers (e.g. we use standard formats for dates to exclude them from the selection, and other formats for percentages to instead include them).
Then, we use the numeral tag of part-of-speech (PoS) tagging module to clean instances where the detected entity is made of several tokens: we select all those marked as numerals (discarding those that are not), and concatenate them when they appear consecutively. 
Finally, we save the extracted numbers along with their offset inside the original text.

\subsection{Unit extraction}

\begin{figure*}
    \centering
    \includegraphics[width=\textwidth]{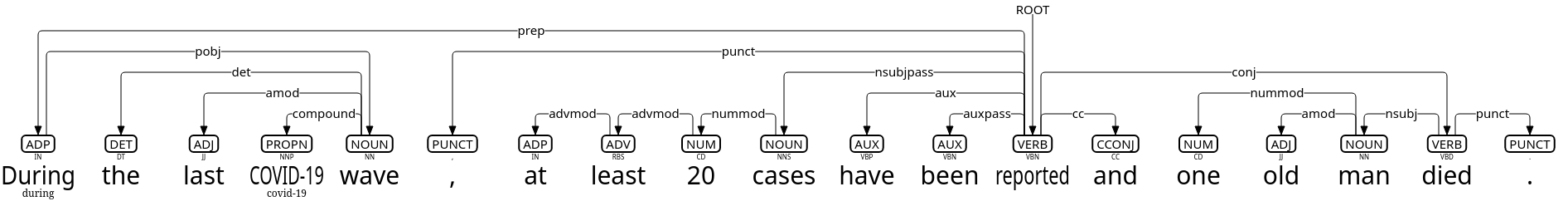}
    \caption{Example of a dependency parsing tree}
    \Description{Example of a dependency parsing tree}
    \label{fig:dep-tree-example}
\end{figure*}

Regarding the unit identification, instead, we want to find them inside the original text by analyzing the relations occurring among each extracted numbers and the other words in the same sentence. 
In order to do so, we make use of the dependency parser and part of speech (PoS) tagging components of our Spacy pipeline.
Our assumption is that the unit of a number consists of nouns, symbols and/or adjectives that could be found close to each extracted number inside the dependency parsing tree (for an example, see Figure \ref{fig:dep-tree-example}).

Given an number extracted earlier, we find the next largest phrase (subtree) spanning its tokens.
We take three approaches.
First, we remove the number tokens and consider the remaining part of the phrase (or the subtree) as a candidate unit. 
In the second approach, we refine this subtree by considering the tokens before and after the number separately. 
We search for the currency tokens in the ones before the number, as it is customary to indicate them before the quantities.
If no such currencies are found, we consider specific part-of-speech tags: \emph{noun}, \emph{proper noun}, \emph{adjective}, and \emph{symbol}.
Specifically, given the list of subtree tokens following the considered number, we look for the longest consecutive sequence of tokens labeled with one of the relevant PoS tags.
If even after this step no candidate unit is found, then we assume that the considered number does not have a unit.
In the third, and last, approach, we apply the same process with an additional step of filtering out all the numbers that do not have a linked predicted unit. 
This step is motivated by the fact that the majority of the numbers will have a linked unit (90\% of annotated numbers had a unit), filtering out those matches that do not will decrease the amount of false positives among the numbers extracted in the first stage and will likely have a positive impact on the overall system's precision.

%
%

\section{Results}

We use the above-mentioned annotated dataset both for the development of the algorithm, and for its evaluation.
In order to do so, we split the dataset into the development set (80\% or 624 excerpts) and test set (20\% or 156 excerpts). 
We evaluate the number and unit extraction separately.
The measures we use come from the information retrieval (IR) domain, treating the spans of text (which could be numbers or [number, unit] combinations) as a kind of ``documents'' of interest in IR.
Specifically, precision is defined as the fraction of spans that are relevant out of all retrieved, recall is the fraction of spans that are retrieved out of all relevant ones, and the F1 score is a harmonic mean of the two measures. 
Moreover, in the cases where both sets of relevant and retrieved spans are empty, we consider the precision, recall, and F1 score equal to 1. Instead, if one of the two sets is not empty, we penalize the performance by setting recall as 0 (when relevant set is empty) and precision as 0 (when retrieved is empty). 
These scores (including F1) are computed per excerpt and then averaged.

\subsection{Number extraction}

We evaluate the number extraction in two ways: a stricter one in which we search for the exactly matching spans between ground and predicted truth, and a ``fuzzy'' one in which we consider a match not only exactly matching spans but also those that overlap for at least a character.
Table \ref{tab:number_extraction} shows the precision, recall, and F1 score for the number extraction on the test set, with two versions: with concatenating consecutive numeral tokens and without. 
Also in fuzzy matching condition, we report the average character distance between the ground truth and the prediction (computed only on those spans that match partially).
As expected, the ``fuzzy'' matching provides higher performance metrics, especially in recall. 
Note that it may be sufficient that the extracted spans are only partially overlapping to be considered as a match if the output of the system will be shown visually to the users who will further modify and enrich it (as is one of the use cases motivating this work).
Further, we find that the step wherein consecutive numerical tokens are combined improves recall, up to 0.87 in strict and 0.97 in fuzzy conditions, but degrades precision. 
Finally, when we consider the average distance in terms of characters between the predicted and ground-truth annotation when they are partially overlapping, we find it to be 6 characters.
Such difference often results in the cases when the number is expressed in words (letters), in which case a mistake of one word could result in many letters (for instance, predicting \emph{``one''} instead of \emph{``one and a half''}).
We choose this version of the number extractor for the next step in order to prioritize the number of numerical matches eligible for unit extraction.


\begin{table*}[t]
\begin{tabular}{l|ccc|cccc} \toprule
                 & \multicolumn{3}{c|}{ Strict } & \multicolumn{3}{c}{Fuzzy} &\\ 
                 & P    & R    & F1   & P    & R    & F1   &  Dist \\ \midrule
Baseline: \texttt{NUM} tagged tokens & 0.42 & 0.60 & 0.48 & 0.64 & 0.94 & 0.73  & 6 chars \\ \midrule
Not concatenate  & 0.83 & 0.80 & 0.79 & 0.88 & 0.84 & 0.83 & 6 chars \\ 
Concatenate      & 0.71 & 0.87 & 0.76 & 0.74 & 0.91 & 0.80 & 6 chars \\ \bottomrule
                    
\end{tabular}
\caption{Precision, recall, and F1 score for the number extraction methods: a baseline and proposed method with two variations: with concatenating consecutive numeral tokens and without. Also in fuzzy matching condition, we report the average character distance between the ground truth and the prediction, for partially overlapping matches.}
\label{tab:number_extraction}
\vspace{-0.4cm}
\end{table*}

\subsection{Unit extraction}

Next, we evaluate the unit extraction module by considering two scenarios. 
First, to evaluate only the performance of the unit extractor, limit our dataset to the numbers which have been correctly extracted in the first step (such that errors in the first step are not propagated to the second), and evaluate the output only on the units of these correctly extracted numbers.
Second, to evaluate the combined performance of both steps, we do not limit the numbers passed on to the unit extractor, and evaluate the final output considering the full set of annotated numbers.
Note that, again, here we present the results on the testing set.

Similarly to the previous step, we consider two versions of the unit extractor. 
Note that here, we report the results of fuzzy matching only. 
In order to improve the precision of our tool, we can use the output of second stage to improve the number extraction. 
Upon examination, we find that the majority of numbers for which no unit is detected by our method are false positives (that is, not extracted correctly in the first step).
Recall that only 10\% of the annotated numbers do not refer to a unit, this is in a stark contrast to the fact that 46\% of extracted numbers we find to be false positive do not have a predicted unit\footnote{This percentage is computed on the development set.}.
Thus, we propose to eliminate such numbers from the final output of the two stages (in the table, dubbed ``Discard numbers w/o predicted units'').

Table \ref{tab:unit_extraction} shows the precision, recall, and F1 score for the unit extraction on the test set.
Note that, because these measures are aggregated per document first, sometimes the aggregated F1 score is not precisely equal to the harmonic mean of aggregated P and R. 
Unlike in the previous task, consider the [number, unit] spans tuple as a kind of ``document'', instead of a single number span.
As expected the results for the first measure are higher because we do not take in account all the false positive predicted numbers (which are not a few, based on the results coming from the evaluation of the first stage).

\begin{table*}[t]
\begin{tabular}{l|ccc|cccc} \toprule
                 & \multicolumn{3}{c|}{ Correct number matches } & \multicolumn{4}{c}{All number matches}  \\ 
                                     & P    & R    & F1   & P    & R    & F1   &  Dist \\ \midrule
Baseline$_1$: BL num + next consecutive token  & 0.51 & 0.52 & 0.51 & 0.32 & 0.48 & 0.37 & 13 chars \\
Baseline$_2$: Final num + next consecutive token     & 0.53 & 0.52 & 0.52 & 0.39 & 0.48 & 0.42 & 13 chars \\ \midrule
CQE \cite{almasian-etal-2023-cqe}     & - & - & - & 0.45 & 0.50 & 0.46 &  11 chars \\ 
CQE \cite{almasian-etal-2023-cqe} \& discard numbers w/o predicted units    & - & - & - & 0.50 & 0.50 & 0.48 & 11 chars \\ \midrule
Number subtree             & 0.75 & 0.75 & 0.75 & 0.57 & 0.69 & 0.60 & 55 chars \\ 
Filter \& keep numbers w/o predicted units    & 0.71 & 0.71 & 0.71 & 0.53 & 0.65 & 0.57 & 14 chars \\ 
Filter \& discard numbers w/o predicted units & 0.77 & 0.68 & 0.71 & 0.63 & 0.62 & 0.60 & 14 chars \\ \bottomrule
\end{tabular}
\caption{Precision, recall, and F1 score (using fuzzy matching) for the unit extraction, with two versions: that keeps or discards the extracted numbers without predicted units. Two baselines and two versions of a number/unit extractor (CQE) are also shown. Two conditions are reported: where only correct number matches are considered (leftmost columns) and where all numbers extracted in the first step are considered (rightmost step). Finally, we report the average character distance between the ground truth and the prediction, for partially overlapping matches.}
\label{tab:unit_extraction}
\vspace{-0.5cm}
\end{table*}

First, we evaluate two baseline approaches wherein (1) a baseline number extractor is used, or (2) the final number extractor is used. 
For both baselines, we consider the next consecutive token (word or a symbol) of the number as its unit. 
We find the approach to have F1 of 0.52 in the more generous condition, which falls to 0.42 when all detected numbers are considered.
When the unit is only partially overlapping, the distance between the detected and ground truth is 13 characters.
We also test the recently proposed quantity extraction framework Comprehensive Quantity Extractor (CQE)~\cite{almasianCQEComprehensiveQuantity2023}, and find that it on average achieves F1 score of just below 0.50.
Note that the tool may need some adjustment to perform well on this task, as the definition of a relevant quantity has been constrained by the experts (for instance, percentages are not of interest for this task), and the library provides changes to the original text that may not be appropriate for this setting.

Instead, we consider the dependency tree of the sentence containing a given number, and find the next largest phrase (subtree) spanning the number's tokens (which involves finding the node pointing to the tokens and taking its subtree).
When we exclude the detected number from this subtree, we may consider it a unit. 
As shown in the table, such an approach usually finds the unit -- note that the performance metrics are computed using fuzzy matching -- however the amount of text selected this way is much larger than necessary, resulting in a distance of 55 characters between the predicted and the actual unit.
Such a high difference in the detected spans (of many words!) makes the above approach nearly useless in the task of pointing out the unit to the system user, as well as for any information extraction.

\begin{figure}
\begin{minipage}[b]{0.95\linewidth}
    \centering
    \includegraphics[width=\textwidth]{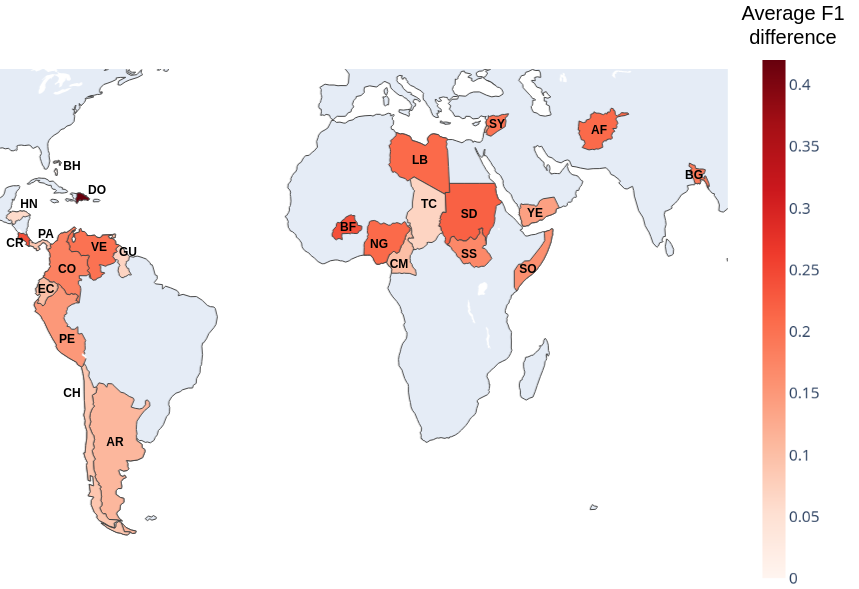}
    \captionof{figure}{Improvement in F1 of the final model vs.~baseline.}
    \Description{Improvement in F1 of the final model vs.~baseline.}
    \label{fig:map_improvement}
\end{minipage}
\hspace{2cm}
\begin{minipage}[b]{0.95\linewidth}
    \centering
    \begin{tabular}{l|rrrrr}\toprule
Sector        & P & R & F1 & dist & docs \\ \midrule 
Shelter         & 0.70 & 0.74 & 0.70 & 15 & 110 \\
WASH            & 0.68 & 0.68 & 0.66 & 18 & 110 \\
Protection      & 0.65 & 0.68 & 0.65 & 16 & 189 \\
Agriculture     & 0.67 & 0.62 & 0.63 & 14 & 41 \\
Education       & 0.64 & 0.67 & 0.63 & 15 & 93 \\
Logistics       & 0.65 & 0.62 & 0.62 & 13 & 47 \\
Health          & 0.62 & 0.64 & 0.61 & 17 & 259 \\
Food security   & 0.62 & 0.59 & 0.58 & 15 & 211 \\
Livelihoods     & 0.62 & 0.58 & 0.57 & 19 & 145 \\
Nutrition       & 0.50 & 0.51 & 0.49 & 17 & 78 \\ \bottomrule
    \end{tabular}
    \captionof{table}{Final model performance by pillar: precision, recall, F1, character distance, and number of documents included.}
    \label{tab:pillars}
\end{minipage}
\end{figure}

Finally, we apply the filters described in Methodology section to allow for a more precise extraction of the unit, resulting in the two last rows in the Table.
We find that by applying this filter, the performance in terms of precision and recall does not degrade substantially, while the distance between the predicted and actual unit decreases to 14 characters.
When we discard the predicted numbers without predicted units, the precision increases while recall decreases, because some numbers without units exist in the annotated set (10\%).
In summary, the final approach attains the best performance in terms of F1 and the amount of character overlap between the predicted and annotated units.

The statistics in the table hide a variability of performance improvements in different geographic locales.
Comparing our final model (with discarding numbers without predicted units) to the baseline$_1$, 
we find that the difference in performance is always non-negative, ranging from 0 improvement for the country with one document that has no numbers (Bahamas) to 0.42 point improvement in case of Dominican Republic.
On average, among all the countries in our datasets, the improvement is 0.16 points. 
Figure \ref{fig:map_improvement} we show the improvement over the baseline per country, showing only countries that have at least 10 documents. 
The improvement in performance is higher for African countries (on average improvement of 0.17) and a bit less for the South American ones (0.15).
This could simply be due to the number of documents available for each country (on average, African countries have 43 documents in our dataset and South American 17 documents).
Further, Table \ref{tab:pillars} shows the performance of the system by pillar (category). 
The system achieves especially high performance on documents related to the humanitarian sectors of \emph{shelter} and \emph{WASH} (\emph{water, sanitation and hygiene}), though further enhancements are needed for the \emph{nutrition}, \emph{food security}, and \emph{livelihoods} categories.
In the future, a more focused study of under-represented locales would be necessary to establish the peculiarities of the content and its impact on the quantity and unit information extraction tasks.

\section{Discussion \& Conclusions}

Here, we introduce a Natural Language Processing (NLP) pipeline designed to extract quantitative information from humanitarian documents. We contribute to the scientific community, by identifying numerical data along with its context, including units and modifiers associated with events, laying the groundwork for creating a data-driven quantity  taxonomy for humanitarian response. 
Although some attempts have been made to create taxonomies of the scientific literature concerning crises and disasters \cite{kuipers2017taxonomy}, little attention has been paid to documents produced during humanitarian action. 
A greater standardization of knowledge in these documents will allow for a more powerful aggregation of statistics and information, supporting tracking and evaluation of efforts.

The quantitative extraction model can be further integrated with spatial \cite{belliardo2023leave}, temporal and organizational information to support humanitarians to quantify the needs, impact of the event and the response per sector and per location. 
The quantitative information extraction from the text can be combined with already available quantitative humanitarian information to enrich the datasets for an efficient response and anticipatory action through predictive models. 
In other words, the outcome of our work can be an input for predictive models to identify the risks and take necessary proactive measures (such as for modeling food insecurity \cite{fiandrino2023impact}).
The annotation data can further be used to pre-train humanitarian based language model to understand and retrieve quantitative information in relation with its unit and event.
Already, efforts have been made \cite{tamagnone2023leveraging} to train a language model specifically for the humanitarian domain, HumBERT, and to systematically evaluate it to measure and mitigate any potential biases that may violate the Leave No One Behind (LNOB) principle.\footnote{\url{https://unsdg.un.org/2030-agenda/universal-values/leave-no-one-behind}}
Such tools may be necessary, as we have shown a great variety in the objects (units) identified in our data, which may need to be grouped semantically instead of linguistically.

The proposed system demonstrates higher precision and recall in extracting quantities and associated units over existing baseline methods, with a clear improvement in specific humanitarian sectors of shelter and water, sanitation, and hygiene (WASH), with potential for improvement in areas including nutrition, food security, and livelihoods. 
Finally, although the system is tailored to humanitarian domain, it can be further specialized to perform well on specific tasks, e.g.~involving people, particular aid provided or needs that remain to be addressed.

\begin{acks} 
DL, KK, and YM acknowledge support from the Lagrange Project of the Institute for Scientific Interchange Foundation (ISI Foundation) funded by Fondazione Cassa di Risparmio di Torino (Fondazione CRT).
DEEP platform is funded by USAID BHA and DFS is the Technical Implementing Partner. 
\end{acks}
\bibliographystyle{ACM-Reference-Format}
\bibliography{hum_num_extraction}

\end{document}